\documentclass[letterpaper]{article} 
\usepackage{aaai2027}  
\usepackage[hyphens]{url}  
\usepackage{graphicx} 
\usepackage{natbib}  
\usepackage{caption} 
\usepackage{algorithm}
\usepackage{amsmath}
\usepackage{booktabs}
\usepackage{multirow} 
\usepackage{makecell} 
\usepackage{graphicx}
\usepackage{array}
\usepackage{array}
\usepackage{colortbl}
\usepackage{algpseudocodex}
\usepackage{amssymb}
\usepackage{xcolor}
\usepackage{tabularx}

\definecolor{movieyellow}{RGB}{252,243,207}
\definecolor{benchpink}{RGB}{232,203,203}
\definecolor{datasetgreen}{RGB}{218,231,209}

\usepackage{newfloat}
\usepackage{listings}
\DeclareCaptionStyle{ruled}{labelfont=normalfont,labelsep=colon,strut=off} 
\floatstyle{ruled}
\newfloat{listing}{tb}{lst}{}
\floatname{listing}{Listing}

\usepackage{booktabs}

\newcommand{\val}[2]{$#1{\scriptscriptstyle\,\pm\,}#2$}
\newcommand{\crcell}[4]{\makecell[l]{C: \val{#1}{#2}\\[1pt] R: \val{#3}{#4}}}

\title{LazySloth: Bounded LLM-based Lazy Tree Search\\ for Fast Long Video Comprehension}

\author {
    Arka Mukherjee\textsuperscript{\rm 1},
    Kaleen Shrestha\textsuperscript{\rm 2},
    Larissa Zhu\textsuperscript{\rm 2},
    Maja Matarić\textsuperscript{\rm 2},
}
\affiliations {
    \textsuperscript{\rm 1}School of Computer Engineering, Kalinga Institute of Industrial Technology (KIIT) Bhubaneswar\\
    \textsuperscript{\rm 2}Computer Science Department, Viterbi School of Engineering, University of Southern California
}

\begin{document}
\nocopyright

\maketitle

\begin{abstract}
Modern vision-language models (VLMs) have shown promising results in long-video understanding due to the rich semantic information they can capture. However, most methods focus on coarse captioning of extracted image frames that are computationally inefficient and require models with large context windows. While past work has explored efficient methods through multimodal retrieval-augmented generation (RAG), they rely on lossy embeddings that lose temporal context and fine-grained detail. Few works to date have investigated how VLM-based query-relevant information retrieval can be optimized. We introduce LazySloth, an efficient tree-based search method that speeds up video comprehension and retrieval tasks 2.9-8.3x (compared to existing agentic methods) through bounded captioning of portions of the video considered irrelevant by a VLM of the video. Compared to contemporary specialized video-understanding VLMs and RAG-based methods, LazySloth achieved similar or better final task accuracy across two recent open-source base VLMs--Gemma 4 31B and Qwen3.6 27B--across four benchmarks. LazySloth reduced the gap between the base open-source model and a closed-source model, GPT-4o. Ablations showed that replacing VLM scene understanding with CLIP-based retrieval cost 8.8–19.9\% in accuracy, while lazy tree construction matches eager construction at a fraction of the captioning cost. With LazySloth, we demonstrate the possibility of faster long-video comprehension without substantial loss in performance. 

\end{abstract}


\section{Introduction}

The frontier for video understanding has moved from short, curated clips~\cite{3666122.3668126} to long-form content such as feature-length films, hour-long lectures, and egocentric recordings~\footnote{\url{https://ego4d-data.org/}}~\cite{Hong_2025_CVPR, Shu_2025_CVPR}, to weeks-long videos spanning multiple topics~\cite{rege-etal-2026-agentic}. Existing work on answering a question from such content focused on explicit \emph{localization} (finding the handful of seconds that matter)~\cite{zuo2025videolucy} and \emph{comprehension} (reasoning over what happens in the localized frames)~\cite{Yeo_2026_CVPR}. Modern VLMs are exceptionally good at processing long-context inputs across modalities~\citep{wang-etal-2025-video}, improving over contrastive methods that showed little success~\cite{10203479}.

\begin{figure}[t]
    \centering
    \includegraphics[width=\columnwidth]{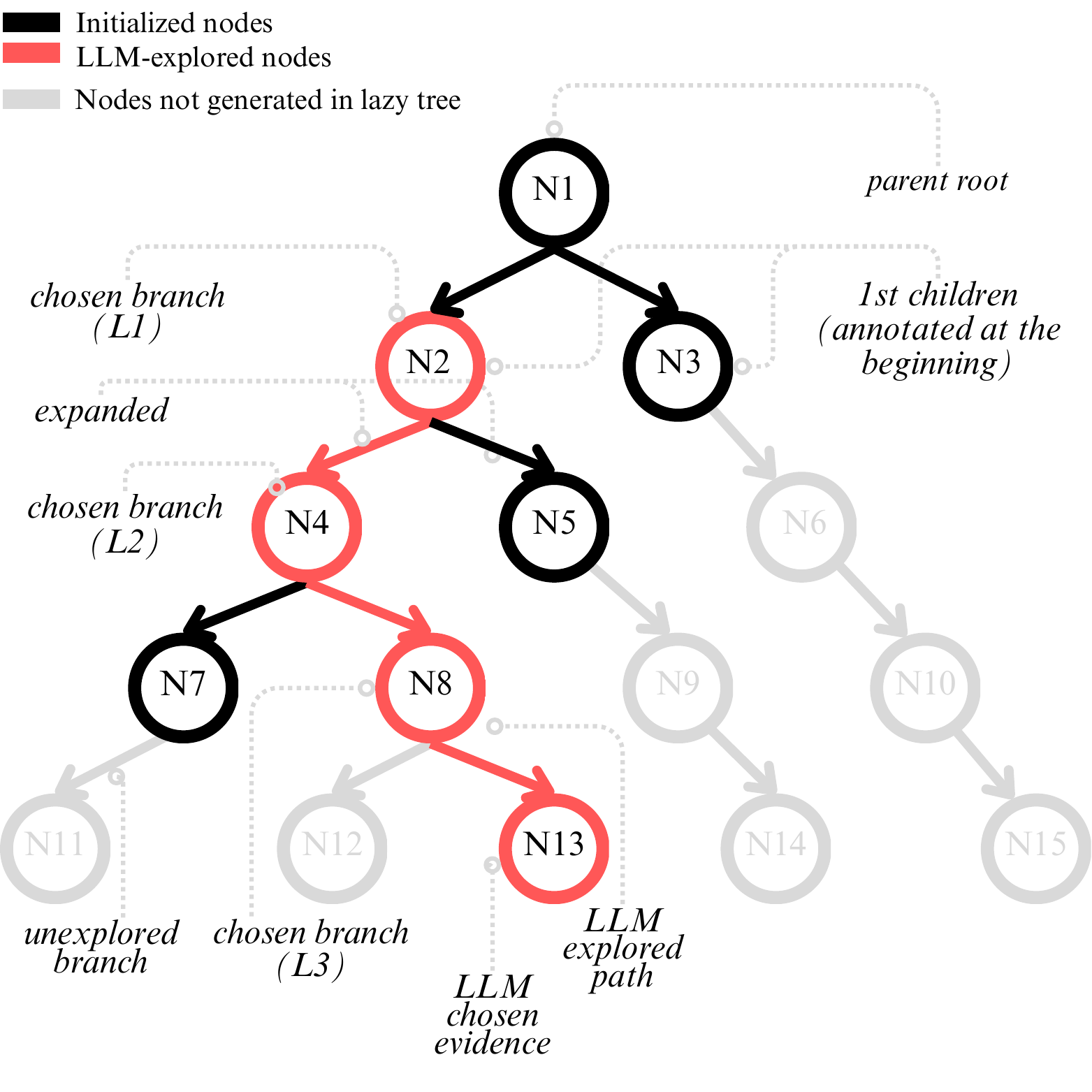}
    \caption{\textbf{Overview of how LazySloth bounds captioning.} Each level captures more fine-grained information, with the parent root representing all of the video. Grayed-out nodes are never visited, which reduces the total time to navigate through the hierarchical tree. }
    \label{fig:lazysloth-method}
\end{figure}

However, the semantic expertise of VLMs comes at a tradeoff, and two main paradigms of VLM-based video understanding have emerged. The first places sampled frames directly into the model's context and asks for an answer in one shot~\cite{shen2025longvu,Shu_2025_CVPR}. This is bounded by the context window and the quadratic cost of attention: an hour of video sampled at even one frame per second (FPS) yields thousands of frames and hundreds of thousands of visual tokens, forcing frame budgets so aggressive that the evidence needed to answer is often not sampled. To combat this, a second paradigm has emerged that converts the video into text: a VLM captions every frame or every fixed-length segment, and an LLM then reasons over the resulting corpus~\cite{zuo2025videolucy,Yeo_2026_CVPR,10.1007/978-3-031-72989-8_4}. This circumvents the context limit and yields an easily searchable artifact. However, this requires one VLM call per unit of video, so the cost scales linearly with duration \emph{regardless of the question asked}. A question about a five-second event still requires captioning the other fifty-nine minutes. Since the VLM call is much more expensive than any other component of the system, this wasted captioning is the bottleneck.

\begin{figure*}[ht]
    \centering
    \includegraphics[width=0.7\textwidth]{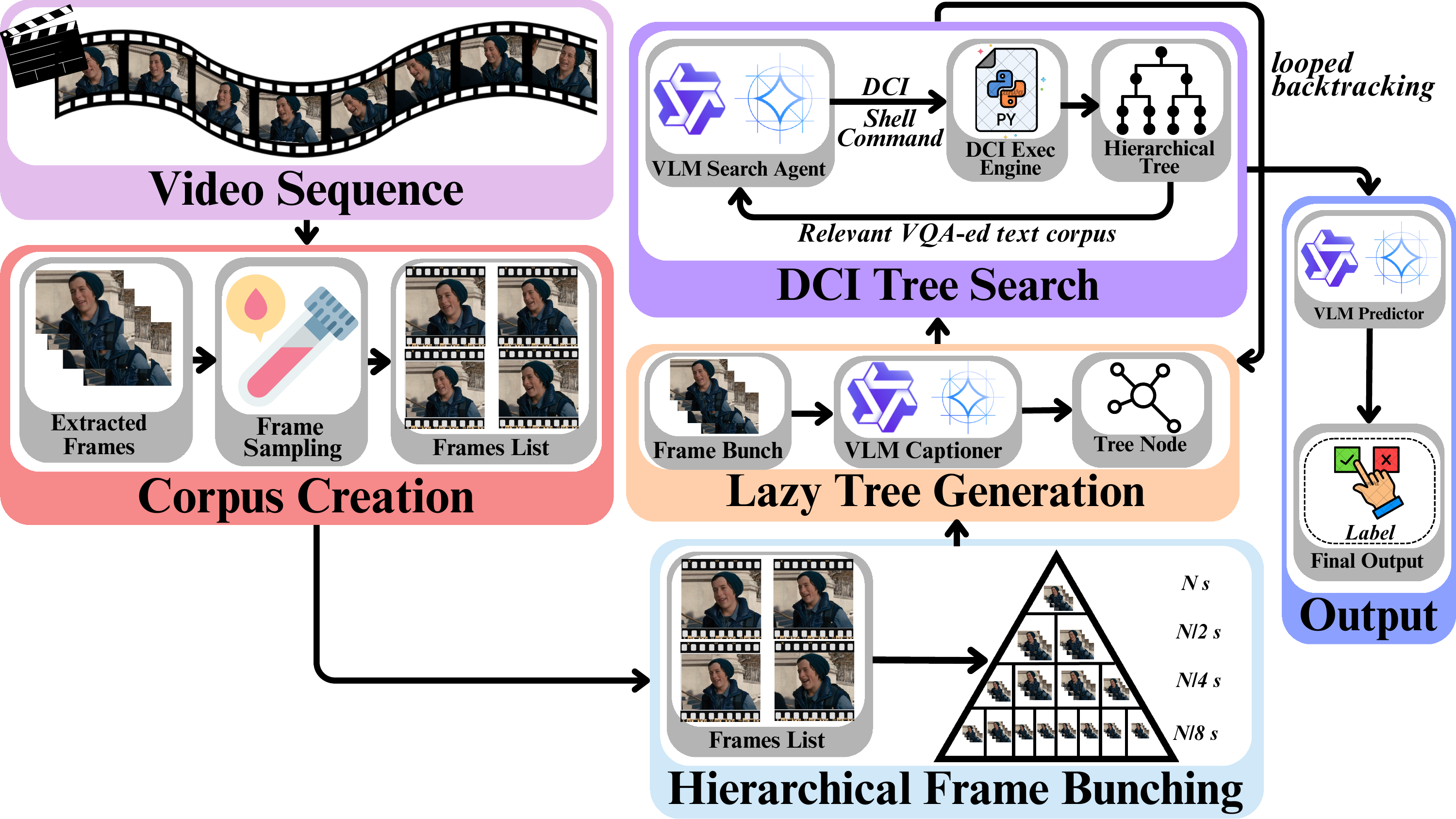}
    \caption{\textbf{Overview of the LazySloth pipeline. (A)} Corpus creation, which produced an ordered list of frames after applying optional uniform frame sampling to the input video. \textbf{(B)} Hierarchical frame bunching, which initialized the first few levels of the search tree. \textbf{(C)} Lazy generation of tree nodes by the captioner VLM $C$. \textbf{(D)} Agentic Direct Corpus Interaction (DCI), where the search VLM $S$ explored the tree with backtracking to traverse alternative branches. \textbf{(E)} Final answer prediction by the reasoner VLM (R), conditioned on the textual evidence and representative frames collected during the search.}
    \label{fig:pipeline}
\end{figure*}

Multimodal retrieval-augmented generation (RAG) addresses precisely this issue by retrieving candidate segments with contrastive image-text encoders before invoking a
VLM \citep{pmlr-v139-radford21a,10.1145/3704268.3742698}. Retrieval is cheap, but the representation is lossy~\cite{yu2025visrag,icaart26} as the contrastive encoder compresses a frame or even a pool of frames into a single vector. This discards temporal, motion, and fine-grained detail. 

Efficient methods retrieve with lossy embeddings while accurate methods caption extensively with VLMs. Few works have asked how \emph{VLM-based} retrieval itself should be optimized. To address this, we introduce \textbf{LazySloth} (Figure~\ref{fig:pipeline}), a tree-based search method that generates captions lazily. LazySloth summarizes the video as a top-down tree that is never fully instantiated. We show a LLM a root summary of the whole clip and a handful of coarse branch summaries, from which it descends into the branch it determines is most relevant to the question. Only that branch's children are captioned, and we repeat this descent until an interval is small enough to expand into per-frame detail. Finally, a multimodal reasoner answers from the evidence surfaced along the path together with the key frames the search resolved to. This bounds the captioning cost to the depth of the descent rather than the length of the video, and grows logarithmically where prior methods grow linearly.

Across four long-video benchmarks and two recent open-source VLMs, LazySloth was 2.9--8.3$\times$ faster than existing agentic methods while matching or exceeding both specialized long-video VLMs and RAG-based pipelines in accuracy. It also narrowed the accuracy gap between the tested open-source models and a closed-source model, GPT-4o.

Our contributions are:
\begin{enumerate}
    \itemsep0em
    \item \textbf{LazySloth}, a lazy hierarchical search method that bounds VLM captioning to the branch the search actually explores, making captioning cost logarithmic rather than linear in video length.
    \item Evidence that \emph{VLM-based} retrieval on frame bunches (that captures temporal change unlike individual frames) can be made \emph{efficient} without falling back on lossy embeddings.
    \item An evaluation across four long-video benchmarks and two open base VLMs against specialized video VLMs, RAG pipelines, and agentic baselines, showing 2.9--8.3$\times$ speedups at comparable or better accuracy.
    \item Ablations that separately isolate the contribution of the hierarchy, of its laziness, and of VLM-based versus embedding-based scene understanding. 
\end{enumerate}

\section{Related Work}

\subsection{Vision–Language Models (VLMs)}
\subsubsection{General Purpose Video VLMs} Open VLM series such as VideoLLaMA~\footnote{\url{https://github.com/damo-nlp-sg/videollama3}}, Qwen-VL~\footnote{\url{https://github.com/qwenlm/qwen3-vl}}, InternVL~\footnote{\url{https://github.com/OpenGVLab/InternVL}}, LLaVA-Video/OneVision~\footnote{\url{https://github.com/EvolvingLMMs-Lab/LLaVA-OneVision-2}}, Gemma~\footnote{\url{https://deepmind.google/models/gemma/gemma-4/}}, and GLM~\footnote{\url{https://z.ai/blog/glm-5}} are extensively used for building agentic video-understanding systems. These VLMs, however, ingest video data as sequences of uniformly sampled frames within their context length budgets, and rely on aggressive truncation or down-sampling for longer and high-resolution videos. Although this allows for graceful scaling, smaller open models degrade after roughly 30 frames~\cite{NEURIPS2024_329ad516}, consistent with the lost-in-the-middle effect~\cite{liu-etal-2024-lost}. Using context-as-memory did not provide usable memory (see experiments in Appendix~\ref{app:fps-sweep}), which motivated our search for efficient retrieval systems that support hundreds of frames.



\subsubsection{Specialized VLMs for Long Videos} Specialized VLMs have also emerged to address the unique problems of long video understanding. VideoLLaMA 3~\cite{zhang2025videoLLaMA3frontiermultimodal}, a well-known model in this space, utilizes high-quality vision-centric image-text datasets in its training corpus and achieved 2-3\% gains on benchmarks like Video Multi-Modal Evaluation (Video-MME)~\cite{video-mme} and Multi-task Long Video Understanding (MLVU)~\cite{11094860}. Other notable models include VideoChat-Flash~\cite{li2025videochatflashhierarchicalcompressionlongcontext} and video-SALMONN 2+~\cite{tang2026videosalmonn}, which employs model architecture and training data innovations to improve video understanding. More recently, VideoChat-R1~\cite{li2025videochatr1enhancingspatiotemporalperception} integrates spatio-temporal specific rewards with reinforcement learning to improve temporal grounding of VLMs. The core limitations of frame sampling and utilizing memory as context, however, remain (Section~\ref{sec:results}).

\subsection{Long Video-Understanding Datasets and Benchmarks}
\label{lit-rev:datasets}

Several high-quality long video-understanding datasets and benchmarks have been released to evaluate foundation model capabilities and train small-scale specialized VLMs. LongVideoBench~\cite{NEURIPS2024_329ad516} and Video-MME~\cite{video-mme} are widely used for model evaluation in a multiple-choice question (MCQ) format, with videos sourced from web platforms such as YouTube\footnote{\url{https://www.youtube.com}}. More recently, evaluation has focused on extremely long (weeks-long, rather than hours) video understanding in VLMs~\cite{11443514}. The Ego4D and EgoLife~\footnote{\url{https://ego4d-data.org/}, \url{https://egolife-ai.github.io/}} corpora have also been utilized for testing long-format egocentric content, with datasets such as EgoSchema~\cite{3666122.3668126} and EgoMemReason~\cite{wang2026egomemreasonmemorydrivenreasoningbenchmark} testing short and week-long comprehension, respectively, in MCQ format. Benchmarks studied in this paper are summarized in Table~\ref{tab:video-datasets}.

\subsection{Methods for Long Video-Understanding}

\subsubsection{RAG-Based Methods} Multimodal retrieval-augmented generation (RAG) has been explored for long-video understanding. \citet{Yeo_2026_CVPR} studied RAG-based methods such as the text-based LightRAG~\cite{guo-etal-2025-lightrag} and HippoRAG~\cite{gutierrez2024hipporag}, and the multimodal Video-RAG~\cite{luo2025videorag}. Our method resolves the concern of relying on lossy multimodal embeddings by utilizing semantic LLM-based scene understanding (Section~\ref{results:ablations}).

\subsubsection{Agentic methods} Several agentic methods have been proposed in the literature to improve VLM-based long video comprehension techniques. Some notable recent work includes VideoLucy~\cite{zuo2025videolucy}, which evaluated the effectiveness of coarse-to-fine-grained memory-based retrieval. Other related work has evaluated the effect of storing specialized memory banks for episodic, semantic, and visual data separately~\cite{Yeo_2026_CVPR}, while~\citet{10.1007/978-3-031-72989-8_4} uses self-reflection-based iterative searching with VLMs. However, these works did not evaluate how VLM-based long video search can be optimized without significant performance reductions.

\section{LazySloth}
\label{sec:ssa-pipeline}

Our proposed method, LazySloth (Figure~\ref{fig:pipeline}), answers a query $Q$ over a long video without captioning every frame. It utilizes three VLMs: a captioner model, $C$, that builds a top-down caption tree lazily, a searcher VLM $S$, and a reasoner VLM $R$. $S$ repeatedly navigates into a branch that it determines to be most relevant. $R$ finally answers the query given the selected caption corpus (temporal context) plus the key frames of the region the search narrowed to (visual evidence). 

After decoding the video clip for frame extraction, we get an ordered frame list $F = \langle f_1 \ldots f_N \rangle$ (timestamps are provided in context). The top-down lazy tree is then initiated, spanning all $N$ frames, which we caption with $C$ on $\leq m_b$ uniformly sampled representative frames. At every step, the root is split into $k$ (branch factor) roughly equal children, each of which represents a contiguous spans of the video. Each span is captioned the same way with $C$. We initialize the evidence corpus as the root caption plus the child captions, and the searcher $S$'s selected key-frame range \texttt{currPos} is set to the whole clip $[0, N)$ at timestamp $t$. Each node is assigned an \texttt{id}.

\begin{algorithm}[h]
\caption{Hierarchical lazy tree generation and retrieval}
\label{alg:main}
\begin{algorithmic}[1]
\Require Video $V$, query $Q$, labels $L$; captioner $C$, searcher $S$, reasoner $R$; branch factor $k$, leaf size $\ell$, key-frame budget $m_r$
\Ensure  answer $\hat{y}$
\State $F \gets \textsc{SampleFrames}(V)$;\quad $N \gets |F|$
\State $C$ captions root $[0,N)$ and its $k$ children
\State $\mathcal{F} \gets \text{children}$
\State $\mathcal{F}_{\text{prev}} \gets \varnothing$
\State $\mathcal{E} \gets \{\text{root, child captions}\}$
\State $[a,b) \gets [0,N)$
\While{round budget remains} \Comment{text-only tree search, budget based on backtracking limit}
  \State $S$ reads $Q$ and the frontier captions $\{\text{caption $C$} \!:\!c \in \mathcal{F}\}$ and emits ONE action:
  \If{\textsc{Drill}\,$C$ \textbf{with} $c \in \mathcal{F} \cup \mathcal{F}_{\text{prev}}$ \textbf{and} $b_c - a_c > \ell$}
     \\ \Comment{$c \in \mathcal{F}_{\text{prev}}\Rightarrow$ backtrack one level}
     \State split $C$ into $k$ children; $C$ captions \emph{only} them 
            \\\Comment{unchosen siblings never captioned}
     \State $\mathcal{F}_{\text{prev}} \gets$ frontier of $C$
     \State $\mathcal{F} \gets \text{children}$
     \State $[a,b) \gets [a_c,b_c)$
     \State $\mathcal{E} \mathrel{+}= \text{their captions}$
  \ElsIf{\textsc{Expand}\,$C$ \textbf{with} $c \in \mathcal{F} \cup \mathcal{F}_{\text{prev}}$ \textbf{and} $b_c - a_c \le \ell$}
     \State C captions each frame of $C$
     \State $[a,b) \gets [a_c,b_c)$
     \State $\mathcal{E} \mathrel{+}= \text{per-frame captions}$
  \ElsIf{\textsc{Answer}\,$y_S$} \Comment{$y_S$: searcher's provisional answer}
     \State \textbf{break} 
  \Else \Comment{if S outputs an invalid id, \textsc{Drill}s on a leaf node, or \textsc{Expand}s on a non-leaf}
     \State prompt $S$ to retry 
  \EndIf
\EndWhile
\State $K \gets$ evenly sample $\le m_r$ frames of $F[a\!:\!b)$ in time order
       \Comment{key frames of the searcher-narrowed region}
\State $y_R \gets R(Q, L, \mathcal{E}, K)$
       \Comment{multimodal: $\mathcal{E}$ = temporal context, $K$ = visual evidence}
\State \Return $\hat{y} \gets y_R$ \textbf{if} $y_R \ne \textsc{Insufficient}$ \textbf{else} $y_S$
\end{algorithmic}
\end{algorithm}

A text-only top-down search then follows, during which $S$ sees the question, an action menu, and a transcript showing the root and current-level captions. Here, searcher $S$ must output exactly one action from the following list:

\begin{enumerate}
    \item \texttt{DRILL <id>}: This action is used to descend into a frontier branch. A node is split into $\leq k$ children, which are then captioned lazily by $C$. For the sibling nodes not chosen by $S$, LazySloth never runs captioning, which bounds the cost. The newly generated children are now the frontier of the tree, \texttt{currPos} is updated, and the new captions are appended to the evidence corpus. We do not discard the previous frontier; instead, we retain it in the evidence corpus so $S$ can \texttt{DRILL} a sibling \texttt{id} to backtrack one level.
    \item \texttt{EXPAND <id>}: If a branch already has $\leq \ell$ frames (i.e, it is at leaf size), the VLM can choose to \texttt{EXPAND} instead. This step captions every frame in that leaf range episodically, which is then appended to the evidence corpus. Finally, \texttt{currPos} is set to that leaf.
    \item \texttt{PROVISIONAL ANSWER <Q>}: The searcher $S$'s answer to $Q$, based on the collected evidence, which ends the search loop.
\end{enumerate}

Finally, both the textual evidence collected from the captioned corpus by the searcher $S$ and $\leq m_r$ representative frames in the leaf nodes of the hierarchical tree that $S$ navigated down to are passed to the multimodal reasoner $R$, which then reasons on a per-frame basis and emits an output label or open-ended text based on the query. The impact of a post-hoc multimodal reasoner is ablated in Table~\ref{tab:reason_vs_agent}.

Let $d = \lceil \log_k (N/\ell) \rceil$ denote the number of levels navigated until a leaf of at most $\ell$ frames is reached. Since each navigation captions at most $k$ new nodes and the final expansion captions at most $\ell$ frames, the total number of captioning operations is bounded by

\begin{equation}
\mathcal{C} \leq dk + \ell
    = k\left\lceil \log_k \frac{N}{\ell} \right\rceil + \ell
    = O\!\left(k \log_k N + \ell\right).
\end{equation}

For a fixed branching factor $k$ and leaf size $\ell$, this simplifies to $O(k \log_k N)$, compared to the $O(N)$ frame-captioning cost incurred by dense captioning. We summarized LazySloth's optimized video search in Algorithm~\ref{alg:main}.

\subsection{Evaluation Benchmarks}
\label{sec:evaluation-benchmarks}

\begin{table}[t]
\centering
\scriptsize
\setlength{\tabcolsep}{3pt}
\begin{tabular}{@{}l cccc@{}}
\toprule
& \makecell{\textit{Video-}\\\textit{MME}}
& \makecell{\textit{LongVideo}\\\textit{Bench}}
& \makecell{\textit{LV}\\\textit{Bench}}
& \makecell{\textit{Ego}\\\textit{Schema}} \\
\midrule
Source        & YouTube & Web & YouTube & Ego4D \\
Domain        & \makecell{6 domains} & \makecell{Multi-topic} & \makecell{Longform} & \makecell{Egocentric} \\
\addlinespace[2pt]
\# Videos        & 900   & 753   & 103   & 500 \\
\# Questions     & 2{,}700 & 1{,}337 & 1{,}549 & 500 \\
\# Options       & 4     & 4--5  & 4     & 5 \\
\addlinespace[2pt]
Avg.\ length (min)   & 17.0 & 7.9 & 67.3 & 3.0 \\
Range (min)   & 0.2--59.7 & 0.1--59.6 & 30.1--139.9 & 3.0 fixed \\
Total duration (h)     & 255.3 & 99.8 & 115.5 & 25.0 \\
\bottomrule
\end{tabular}
\caption{Long-video comprehension benchmarks used in our evaluation:
Video-MME \citep{video-mme}, LongVideoBench \citep{NEURIPS2024_329ad516},
LVBench \citep{11443514}, and EgoSchema \citep{3666122.3668126}.
All four are multiple-choice and scored by exact match. The suite spans web, film/television, and egocentric content, and in total, we evaluate almost 500 hours of video.}
\label{tab:video-datasets}
\end{table}

We evaluated the proposed framework on four established MCQ video question-answering benchmarks with well-maintained leader boards, spanning movie clips, general web videos, long-form videos, and egocentric videos, following the discussion in Section~\ref{lit-rev:datasets}: Video-MME~\cite{video-mme}, LongVideoBench~\cite{NEURIPS2024_329ad516}, LVBench~\cite{11443514}, and EgoSchema~\cite{3666122.3668126}. Table~\ref{tab:video-datasets} summarizes each dataset's statistics. For all of these benchmarks, we conducted a full run and reported task accuracy as an exact match of the LLM's output MCQ label to the ground truth.


\subsection{Baselines}
For each dataset, we first quantified random baseline performance as $ 1/N$, where $N$ is the number of MCQ options for Video-MME, LongVideoBench, LVBench, and EgoSchema. 

We compared LazySloth to multiple classes of methods starting with general video-understanding models, specifically Qwen3.6 27B\footnote{\url{https://huggingface.co/Qwen/Qwen3.6-27B}} and Gemma 4 31B \footnote{\url{https://huggingface.co/google/gemma-4-31B-it}}, using at most 64 frames sampled from each video to stay within their context windows~\cite{yin-mmlm-survey,video-mme,doorenbos2025video}. Next, specialized video-understanding models such as VideoChat-R1 \cite{li2025videochatr1enhancingspatiotemporalperception} and VideoLLaMA 3 \cite{zhang2025videoLLaMA3frontiermultimodal} were tested. To understand how our method compares to RAG-based methods, we compared it against the text-based LightRAG~\cite{guo-etal-2025-lightrag} followed by the multimodal Video-RAG~\cite{luo2025videorag}. Finally, among modern agentic baselines, we included VideoLucy~\cite{zuo2025videolucy} and WorldMM~\cite{Yeo_2026_CVPR}. 


\subsection{Ablations}
\label{method:ablations}

We designed two ablations to investigate (a) the behavior of lazy tree generation, (b) the impact of the tree data structure, and (c) the effectiveness of using VLMs instead of multimodal embeddings. We replaced the searcher VLM $S$ with embedding-based retrieval by scoring the tree nodes using Contrastive Language–Image Pretraining Score (CLIPScore)~\cite{hessel-etal-2021-clipscore}. This enabled greedy best-first search (Greedy BFS) over the tree.

The searcher $S$ first navigated that tree lazily with \texttt{DRILL} and \texttt{EXPAND}, and generated its answer using the collected evidence. This allowed us to illustrate the impact of \textbf{VLM-based scene understanding}. Next, we ablated the effect of lazy tree generation by eagerly constructing the full hierarchy with CLIPScore-based nodes. The searcher $S$ then received the full tree as context while collecting evidence. We kept all other components, including the backtracking from the main method, intact, ablating the \textbf{lazy, on-demand (top-down, implicitly pruned)} tree construction of LazySloth. These comparisons allowed us to illustrate the efficiency gains of Algorithm~\ref{alg:main} over naive brute-forced methods.


\subsection{Experimental Setup \& Models Tested}

All experiments on existing agentic methods and LazySloth were repeated on two base searcher $S$ and reasoner $R$ VLMs (Qwen3.6 27B and Gemma 4 31B). We ran Qwen 3.6 and Gemma 4's compiled GGML Universal File (GGUF) locally through LM Studio\footnote{\url{https://lmstudio.ai/}}. The captioner model $C$ was set to Nvidia's Nemotron 3 Nano Omni 30B for an equitable comparison of each method's merit (which was chosen based on caption quality experiments outlined in Appendix~\ref{app:caption-quality}). We loaded Nemotron 3's compiled GGUF locally through LM Studio. All experiments were run on four Nvidia H100 80 GB, one RTX Pro 6000 96 GB, and two RTX 3090 24 GB GPUs. All videos were sampled at 1 FPS and models initialized with a context window of 79,104 tokens to fit on our available hardware. 


\begin{table}[t]
\centering
\scriptsize
\setlength{\tabcolsep}{3.5pt}
\begin{tabular}{@{}l cccc@{}}
\toprule
 & \multicolumn{4}{c}{\textbf{Accuracy (\%)}} \\
\cmidrule(l){2-5}
\textbf{Method} &
\makecell{Video-MME\\(w/o subs)} & \makecell{LongVideo\\Bench} &
\makecell{LV\\Bench} & \makecell{Ego\\Schema} \\
\midrule
Random baseline & 25.0 & 21.3 & 25.0 & 20.0 \\
\midrule
\multicolumn{5}{@{}l}{\textit{Direct inference (64 frames)}} \\
\;\;GPT-4o$^{\ddag}$      & 71.9 & 66.7 & 48.9 & 72.2 \\
\;\;Qwen 3.6 27B          & 45.7 & 31.8 & 27.7 & 38.4 \\
\;\;Gemma 4 31B           & 65.8 & 52.8 & 40.7 & 65.4 \\
\midrule
\multicolumn{5}{@{}l}{\textit{Specialized video models}} \\
\;\;VideoChat-R1.5        & 52.5 & 49.8 & 33.4 & 41.6 \\
\;\;VideoChat-Flash$^\dag$ & 55.6 & --- & 33.2 & 44.1 \\
\;\;VideoLLaMA 3          & 56.2 & 51.0 & 35.4 & 60.6 \\
\;\;Time-R1$^\dag$         & ---  & ---  & 37.6 & 31.1 \\
\midrule
\multicolumn{5}{@{}l}{\textit{RAG-based}} \\
\;\;LightRAG$^\dag$        & 46.6 & ---  & 30.4 & --- \\
\;\;Video-RAG 7B$^\dag$    & 55.4 & ---  & 33.1 & --- \\
\midrule
\multicolumn{5}{@{}l}{\textit{Agentic}} \\
\;\;VideoLucy (Qwen)      & 34.3 & 43.9 & 35.4 & 51.1 \\
\;\;VideoLucy (Gemma)     & \textbf{71.3}  & \textbf{65.7} & \textbf{53.8} & \underline{70.9} \\
\;\;WorldMM (Qwen)      & --- & --- & 29.5 & --- \\
\rowcolor{cyan!10}
\;\;\textbf{LazySloth} (Qwen)  & 61.4 & 53.2 & 39.1 & \textbf{73.9} \\
\rowcolor{cyan!10}
\;\;\textbf{LazySloth} (Gemma) & \underline{66.5} & \underline{59.5} & \underline{45.9} & 69.1 \\
\bottomrule
\end{tabular}
\caption{\textbf{Accuracy results on four long video-understanding benchmarks.} 
\textbf{Bold} marks the best open-source result per benchmark; \underline{underline} marks
the second-best open-source result.
\dag~Numbers sourced from~\citet{Yeo_2026_CVPR}.
\ddag~GPT-4o numbers are taken from the public leaderboards of the respective
benchmarks; we do not report a newer GPT release, as none could be located on
the maintained leaderboards.}
\label{tab:main-results}
\end{table}

\section{Results}\label{sec:results}

Table~\ref{tab:main-results} shows the accuracy of LazySloth compared to existing models and methods. Our method outperformed specialized models and retrieval-based methods. On Qwen 3.6 27B, LazySloth exceeded the strongest specialized video-understanding model  on every benchmark: by 5.2 points over VideoLLaMA~3 on Video-MME, 2.2 on LongVideoBench, 1.5 over Time-R1 on
LVBench, and 13.3 over VideoLLaMA~3 on EgoSchema. Importantly, these gains were realized with a general-purpose base model (i.e., backbone), and no video-specific training. LazySloth also outperformed the best RAG-based baseline (Video-RAG 7B) by at least 6.0 points on the two benchmarks where their numbers were available, which was consistent with our claim that embedding retrieval discards fine-grained detail required for sophisticated video understanding.

Among other agentic methods such as VideoLucy, using Qwen3.6 27B as the backbone, we saw that LazySloth improved over VideoLucy by 27.1, 9.3, 3.7, and 22.8 points, and over WorldMM by 9.6 points on LVBench. The margin over direct inference was 11.4--35.5 points. We note that VideoLucy underperformed compared to direct inference on Video-MME with the same backbone (34.3\% vs.\ 45.7\%), whereas LazySloth improved on it by 15.7\%.

Improvements were markedly smaller on Gemma 4 31B, where LazySloth trailed VideoLucy on all four benchmarks by 1.8--7.9 points, which demonstrated that the improvement margin was largely dependent on the specific model: on average, Qwen3.6 27B gained 21.0\% in accuracy with our pipeline compared to 4.1\% for Gemma 4 31B over direct inference. Further experiments with more backbone VLMs would confirm this pattern as a claim. Despite lower performance gains with Gemma 4, LazySloth achieved comparable accuracies at 2.9--8.3$\times$ lower median wall-clock time per video than competing methods, as we discuss next.

\subsection{Efficiency Analysis}~\label{sec:eff-analysis}


\begin{figure}[t]
    \centering
    \includegraphics[width=0.9\columnwidth]{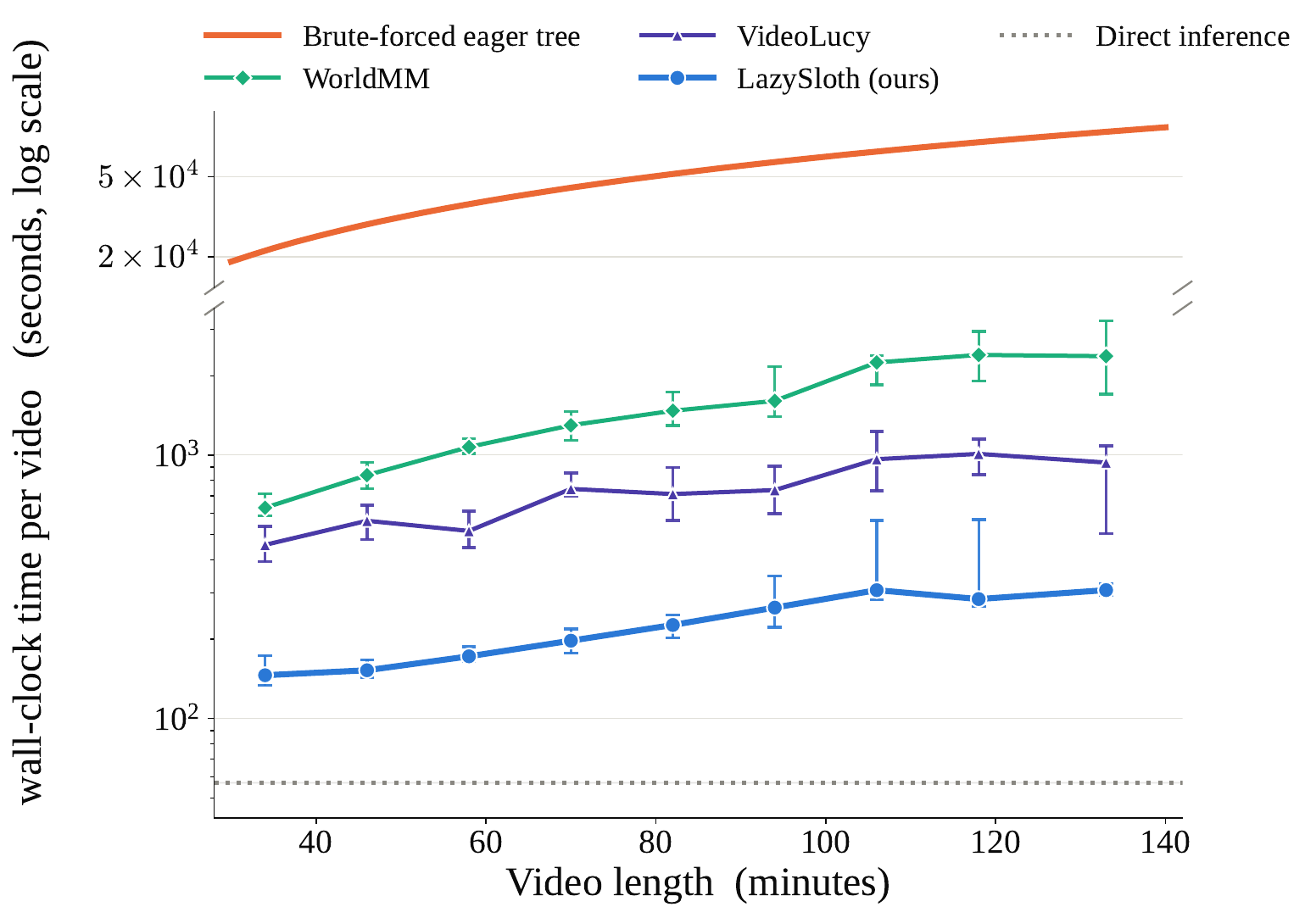}
    \caption{Efficiency comparison of LazySloth, a brute-forced eager tree, VideoLucy, and WorldMM by median seconds per video (log y, broken above the working band) on LVBench, which was chosen as it has the longest videos in our test corpus. Bootstrapped 95\% confidence intervals are plotted at each video length value. All methods shared the Qwen3.6-27B reasoner and Nemotron captioner.}
    \label{fig:efficiency}
\end{figure}

To better understand the gains realized with LazySloth, we first compared the median wall clock time across our method and the two agentic baselines, VideoLucy and WorldMM. As a control, we included the brute-forced eager bottom-up tree (Section~\ref{method:ablations}). For these tests, we focused only on the LVBench dataset as it contained the longest videos in our test set (Table~\ref{tab:video-datasets}). Figure~\ref{fig:efficiency} shows that our method was the fastest, with median time to answer a single LVBench query at 194 seconds (secs) compared to 611 secs on VideoLucy, and 1,139 secs on WorldMM. On the longest videos (140 minutes long), our method was 3$\times$ faster than VideoLucy and 7.7$\times$ faster than WorldMM. The gap with the brute-forced eager bottom-up tree remained disproportionately high at 287$\times$ slower than LazySloth. Direct inference, which feeds 64 frames into the VLM's memory, stayed constant at 57 secs, and resulted in notably lower task accuracies. Additional discussion about efficiency can be found in Appendix~\ref{app:efficiency-analysis}.

\begin{figure}[ht]
    \centering
    \includegraphics[width=0.8\columnwidth]{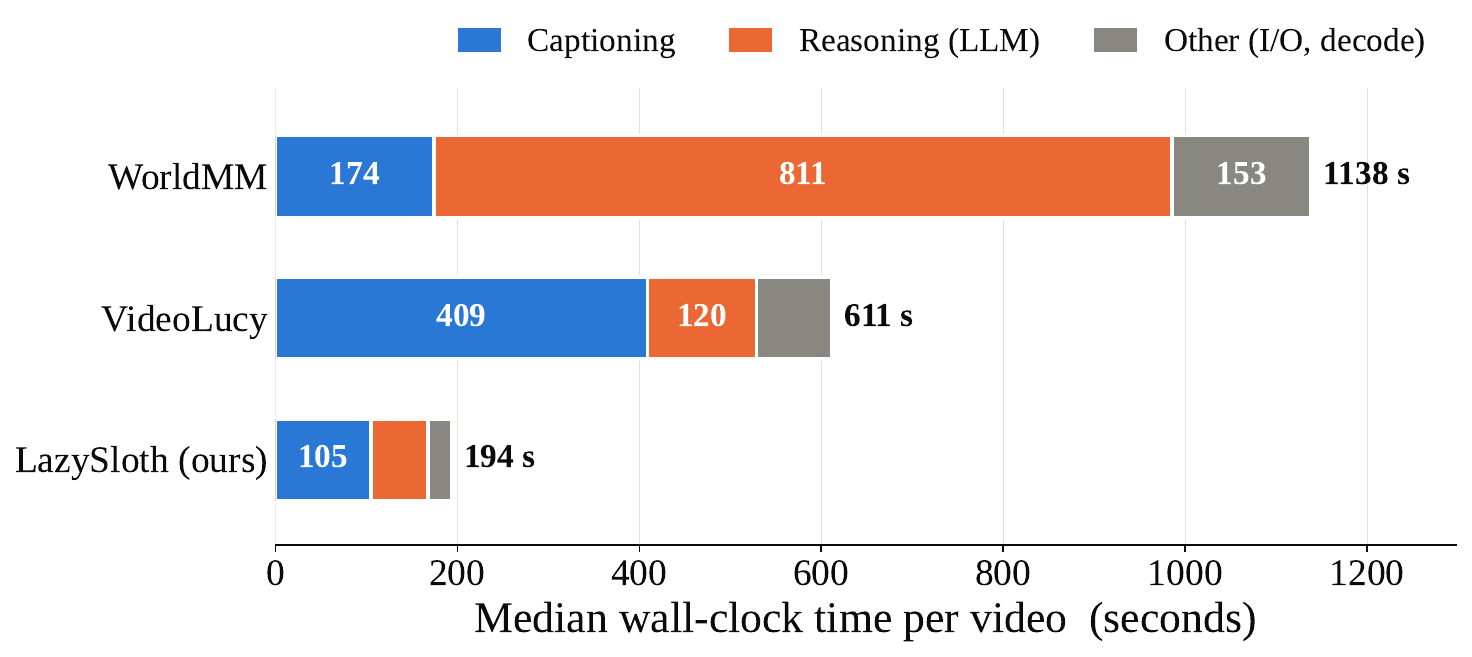}
    \caption{WorldMM vs VideoLucy vs LazySloth median wall clock time per video on LVBench decomposition by captioning, reasoning, and other (I/O, etc.). VideoLucy spends a lot of time captioning; WorldMM spends a lot on reasoning. Our method optimizes both frontiers for the lowest total time.}
    \label{fig:decomposition}
\end{figure}

Next, we decomposed each method into three categories: time spent captioning, time spent reasoning, and overhead (e.g., video decode, I/O, etc). Figure~\ref{fig:decomposition} shows the results on LVBench. We observed that reasoning dominated WorldMM's median wall clock time (811s), as it used triple-extraction with VLMs across episodic, semantic, and visual memories. VideoLucy, on the other hand, was dominated by its multi-frame coarse-window captions (67\%) as compared to reasoning. Our method optimized both frontiers, and neither captioning nor reasoning dominated, ensuring 2.9--8.3$\times$ faster median wall clock time than the baselines.

\subsection{Behavioral Analysis}

\begin{figure}[t]
    \centering
    \includegraphics[width=0.8\columnwidth]{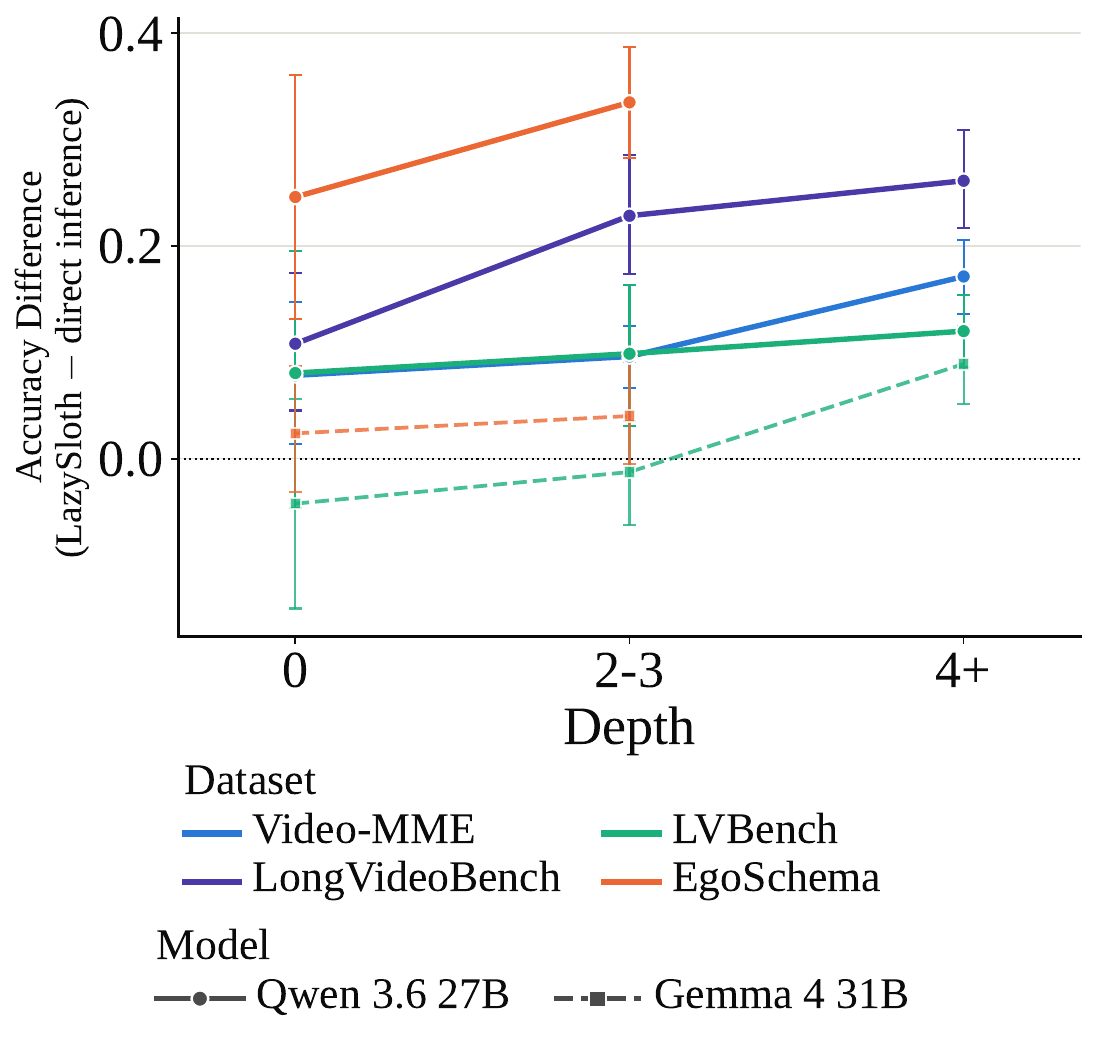}
    \caption{Comparing the accuracy difference between LazySloth and direct inference across benchmarks and models with respect to search tree depth.}
    \label{fig:behavior-depth}
\end{figure}

\begin{figure}[t]
    \centering
    \includegraphics[width=0.7\columnwidth]{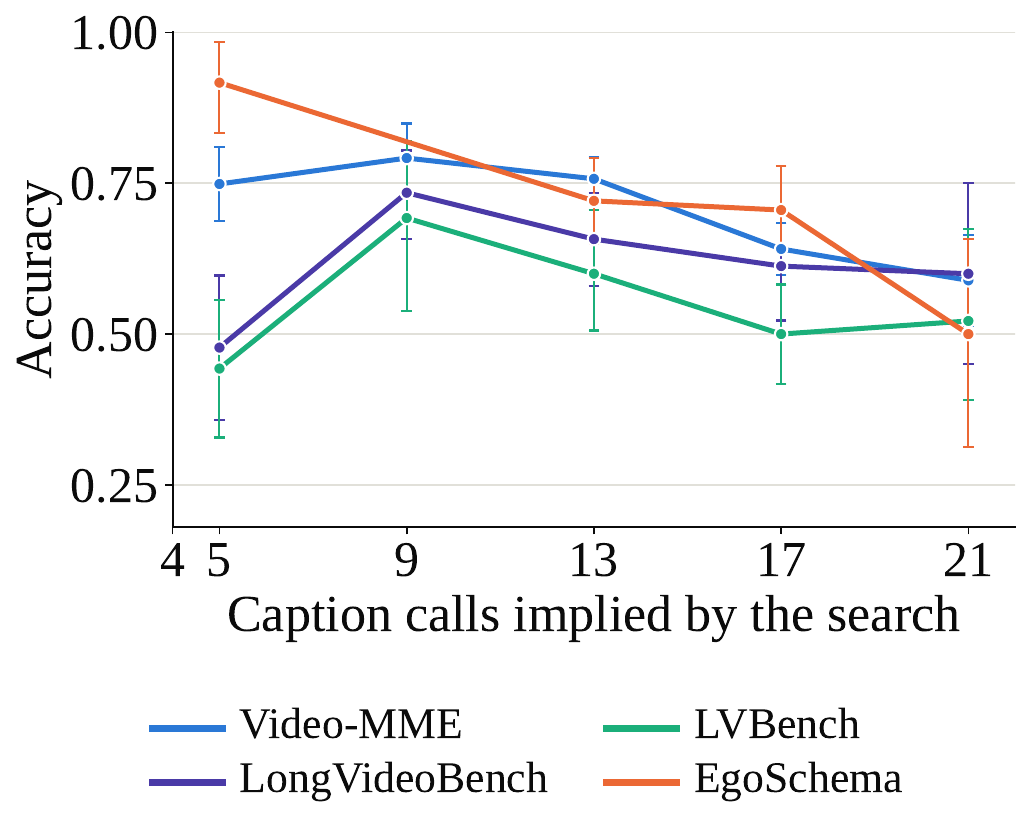}
    \caption{Accuracy with respect to the caption budget used by the tree search with Qwen3.6 27B, where $budget = 1$ \text{root} $+$ $4$ \text{seed branches} $+$ $4$ \text{per successful} \texttt{DRILL} $+$ $4$ \text{per} \texttt{EXPAND}. Beyond a single descent into the tree, accuracy declined across all benchmarks.}
    \label{fig:behavior-budget}
\end{figure}

\begin{figure}[t]
    \centering
    \includegraphics[width=0.8\columnwidth]{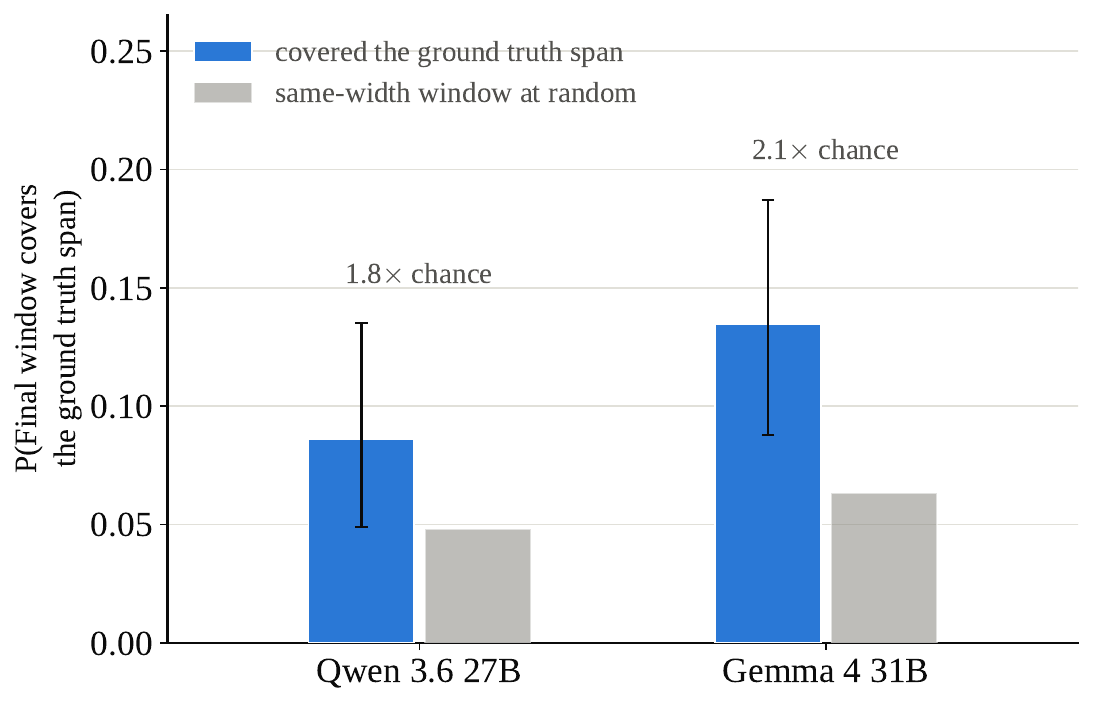}
    \caption{Analysis of temporal localization on correctly answered LongVideoBench questions that were annotated with specific timestamps. Blue: the portion where LazySloth's final window contained the referenced span. Gray: the rate expected from placing a window of the same width at random, averaged per question.}
    \label{fig:behavior-localization}
\end{figure}


We next characterized how LazySloth's search behaves and what its behavior predicts about the final accuracy. Figure \ref{fig:behavior-depth} plots the accuracy difference between LazySloth and direct inference against the depth the search reached. On Qwen3.6 27B, we noticed positive differences across all tested benchmarks, while on Gemma 4 31B, we noticed the differences were zero throughout. This was consistent with the model-based performance differences in Section~\ref{sec:results}. Multi-level tree search did not impact Gemma 4 31B's final accuracy, and likely negatively hurt its long video capabilities.


Next, Figure~\ref{fig:behavior-budget} analyzes the performance of LazySloth with respect to the caption budget during search on Qwen 3.6 27B. The first descent into the tree (moving from five caption calls to nine) clearly helped the VLM, as accuracy rose by 25\%. Beyond that, accuracy fell on Video-MME, LongVideoBench, and LVBench. On EgoSchema, accuracy declined monotonically, likely as videos in the dataset were extremely short (three minutes each). This decline was largely correlated with harder questions, where the search failed to resolve the query cheaply in higher levels of the tree. 

We additionally investigated LazySloth's ability to find the exact span in the video that contains the answer to the query. Compared to randomly choosing similar-sized windows, Figure \ref{fig:behavior-localization} shows each VLM significantly improved ($1.8-2.1\times$) at identifying the correct window (Qwen3.6 27B $n=163$ and Gemma 4 31B $n=171$). However, while the choice was not arbitrary, the final window did not contain the actual span for $86-91\%$ of the questions, where the model still answered the question correctly.

\subsection{Ablation Results}
\label{results:ablations}
We replaced VLM captioning with a frozen CLIP encoder, and found that accuracy fell by $8.8$--$19.9$\% with CLIPScored nodes compared to a Qwen3.6 27B base model (Table \ref{tab:ablation}). VLM-based scene understanding was a major contributor to LazySloth's performance. Holding the scorer fixed at CLIPScore, the lazy top-down and eager bottom-up construction differed by at most $3.1$\% with no consistent direction (Video-MME $41.5$ vs.\ $43.4$, LongVideoBench $40.5$ vs.\ $42.1$, EgoSchema $55.6$ vs.\ $56.4$). Lazy expansion largely did not affect accuracy, explaining how LazySloth speeds up video retrieval~(Figure~\ref{fig:efficiency}) without a massive performance loss. 

Finally, we explored the effect of having a separate reasoner VLM in LazySloth, and found that the search trajectory largely determined the final answer (Table \ref{tab:reason_vs_agent}). The reasoner improved micro-averaged accuracy by $0.3$\%, with no clear directionality. The searcher's provisional answer was thus very nearly as accurate as the final one, and removing the reasoner's VLM call could improve efficiency further.

\begin{table}[t]
\centering
\scriptsize
\setlength{\tabcolsep}{3pt}
\begin{tabular}{@{}l cccc@{}}
\toprule
 & \multicolumn{4}{c}{\textbf{Accuracy (\%)}} \\
\cmidrule(l){2-5}
\textbf{Configuration} &
\makecell{Video-MME} & \makecell{LongVideoBench} &
\makecell{LVBench} & \makecell{EgoSchema} \\
\midrule
\textbf{LazySloth} & 61.4 & 53.2 & 39.1 & 73.9 \\
\midrule
\addlinespace[2pt]
\multicolumn{5}{@{}l}{\textit{Scene understanding (CLIPScore, greedy best-first)}} \\
\;\;lazy top-down tree        & 41.5 & 40.5 & 27.2
 & 55.6 \\
\;\;eager bottom-up tree      & 43.4 & 42.1 & 30.3 & 56.4 \\
\bottomrule
\end{tabular}
\caption{Ablation study of LazySloth with Qwen3.6 27B. Group (a) varies the memory structure while keeping
VLM-based scene understanding; group (b) replaces VLM captioning and search with a frozen
CLIP encoder that scores each tree node by CLIPScore and navigates by greedy BFS.}
\label{tab:ablation}
\end{table}

\begin{table}[t]
\centering
\scriptsize
\setlength{\tabcolsep}{5pt}
\renewcommand{\arraystretch}{1.05}
\begin{tabular}{lrrr}
\toprule
\textbf{Dataset} &
\makecell[c]{\textbf{Searcher (\%)}} &
\makecell[c]{\textbf{Reasoner (\%)}} &
\makecell[c]{\textbf{$\Delta$ (Reasoner$-$Searcher)}} \\
\midrule
EgoSchema       & 76.6 & 76.9 & +0.3 \\
LongVideoBench  & 58.6 & 60.8 & +2.2 \\
LVBench         & 42.5 & 41.4 & $-1.1$ \\
Video-MME       & 62.3 & 62.6 & +0.2 \\
\midrule
\textbf{MCQ Micro-Avg} &
\textbf{58.0} &
\textbf{58.3} &
\textbf{+0.3} \\
\bottomrule
\end{tabular}

\caption{Agreement between the searcher's selected answer and the reasoner's final answer with Qwen 3.6 27B. We observe that using a separate reasoner results in a 0.3\% accuracy gain over always trusting the search model's answer.}
\label{tab:reason_vs_agent}
\end{table}












\section{Conclusion}
We introduce LazySloth, a tree-based search method for long-video understanding that bounds captioning to regions a VLM deems relevant rather than exhaustive captioning. Across four benchmarks and two open-source VLMs, LazySloth was $2.9$ -- $8.3\times$ faster than existing methods while matching or exceeding specialized video-understanding VLMs and RAG-based methods in accuracy. Detailed failure analysis showed gaps in VLM behavior where it navigated to the wrong window or degraded in performance with extra caption calls. Further, our ablations show LazySloth remained competitive due to VLM-based scene understanding, and lazy construction was close in accuracy with eager construction, thereby resulting in a method that speeds up long-video retrieval without a substantial cost in task performance. 

\section{Limitations}

One of the empirical shortcomings of LazySloth is that we observed our method assumed that long-video comprehension can be reliably localized to subtrees, as we observed poor backtracking and navigation to other branches. Our method also relied on the underlying caption quality generated across all resolutions, which we quantify in Appendix~\ref{app:caption-quality}. Additionally, our tests remain limited to two base models, without uncertainty quantification with multiple runs. 


\bibliography{aaai2027}

\clearpage
\appendix
\section*{Appendices}



\section{Symbol Glossary}~\label{app:glossary}

Table~\ref{tab:notation} presents a glossary of all formalization symbols used in Section~\ref{sec:ssa-pipeline} and Algorithm~\ref{alg:main}. 

\section{FPS Sweep Results}~\label{app:fps-sweep}

Table~\ref{tab:acc_only} showcases uniform FPS sampling results on a few short video datasets: DFEW~\cite{jiang2020dfew}, CCDb-IB~\cite{10896249}, CCDb-HG~\cite{10581954}, CCDb+, SEMPI~\cite{10.1145/3678957.3685752}, and RoomReader~\cite{reverdy-etal-2022-roomreader}. We specifically selected them as they are a few seconds long and therefore fit entirely into VLM context windows at high FPS sampling rates. 

The sweep showed that adding frames did not add usable information. Neither model improved monotonically as sampling rose from 1 FPS to every frame. Qwen 2.5 VL 7B\footnote{\url{https://huggingface.co/Qwen/Qwen2.5-VL-7B-Instruct}} on RoomReader fell from 47.0\% at 1 FPS to 11.5\% when given the full frame set, and on SEMPI from 54.0\% to 32.7\%. Gemma 4 E4B\footnote{\url{https://huggingface.co/google/gemma-4-E4B}}, on the other hand, stayed essentially flat across all four rates.

Since these clips are short enough to fit entirely in context, the ceiling is not the context window but the model's ability to use what is already in it. This experiment further motivated a retrieval-based design. 

\begin{table*}[t]
\centering
\scriptsize
\setlength{\tabcolsep}{4pt}
\renewcommand{\arraystretch}{1.1}
\begin{tabular}{@{}ll cccc}
\toprule
\textbf{Tier} & \textbf{Dataset} & \textbf{``All'' (31\,FPS)} & \textbf{24\,FPS} & \textbf{12\,FPS} & \textbf{1\,FPS} \\
\midrule
\multicolumn{6}{@{}l}{\textit{Qwen 2.5 VL 7B} (Unsloth bf16 GGUF)}\\
 DFEW    & \crcell{26.7}{22.7}{12.3}{6.7}  & \crcell{27.4}{22.4}{14.7}{7.4}  & \crcell{32.2}{24.0}{26.2}{13.7} & \crcell{32.5}{22.7}{30.9}{21.8} & \val{14.05}{2.51} \\
 CCDb-IB & \crcell{23.5}{10.1}{18.0}{8.4}  & \crcell{28.2}{8.4}{21.7}{7.9}   & \crcell{22.6}{8.2}{22.2}{7.0}   & \crcell{27.2}{8.0}{14.8}{6.7}   & \val{12.63}{2.39} \\
 CCDb-HG & \crcell{16.7}{22.5}{7.4}{10.0}  & \crcell{16.2}{21.5}{10.1}{13.5} & \crcell{16.8}{20.4}{13.9}{12.9} & \crcell{20.1}{23.8}{17.8}{13.7} & \val{16.73}{2.74} \\
 CCDb+   & \crcell{54.4}{20.7}{20.1}{12.1} & \crcell{59.8}{25.5}{22.7}{12.1} & \crcell{60.7}{25.1}{30.2}{12.2} & \crcell{56.7}{19.9}{57.5}{10.5} & \val{49.87}{3.35} \\
 SEMPI      & \crcell{32.7}{15.4}{1.2}{2.6}   & \crcell{44.8}{18.3}{3.4}{3.9}   & \crcell{43.2}{17.7}{35.4}{11.3} & \crcell{54.0}{19.5}{56.3}{15.4} & \val{50.07}{3.62} \\
 RoomReader & \crcell{11.5}{7.0}{4.0}{5.0}    & \crcell{32.8}{17.1}{21.9}{10.9} & \crcell{34.1}{21.2}{29.3}{18.0} & \crcell{47.0}{38.5}{56.0}{10.0} & \val{49.86}{3.45} \\
\midrule
\multicolumn{6}{@{}l}{\textit{Gemma 4 E4B} (Unsloth bf16 GGUF)}\\
\midrule
 DFEW    & \crcell{30.8}{26.7}{36.2}{23.1} & \crcell{31.3}{25.9}{35.6}{22.9} & \crcell{30.9}{25.3}{38.3}{25.4} & \crcell{32.8}{25.0}{38.1}{20.3} & \val{14.05}{2.51} \\
 CCDb-IB & \crcell{25.3}{11.4}{26.4}{9.6}  & \crcell{26.4}{12.3}{26.6}{9.2}  & \crcell{26.1}{10.5}{27.5}{9.9}  & \crcell{24.2}{9.6}{26.8}{8.5}   & \val{12.63}{2.39} \\
 CCDb-HG & \crcell{16.3}{23.5}{21.1}{22.7} & \crcell{18.6}{24.2}{22.8}{24.2} & \crcell{18.0}{23.5}{20.9}{23.0} & \crcell{18.4}{23.7}{16.2}{14.2} & \val{16.73}{2.74} \\
 CCDb+   & \crcell{63.0}{25.7}{8.1}{5.6}   & \crcell{65.8}{26.4}{9.7}{7.6}   & \crcell{63.8}{23.9}{10.1}{7.1}  & \crcell{63.5}{27.0}{7.4}{5.5}   & \val{49.87}{3.35} \\
 SEMPI      & \crcell{65.0}{12.0}{61.0}{12.0} & \crcell{68.1}{9.8}{67.5}{12.9}  & \crcell{68.5}{13.2}{71.3}{12.0} & \crcell{61.2}{17.5}{64.5}{12.5} & \val{50.07}{3.62} \\
 RoomReader & \crcell{36.9}{23.8}{39.6}{27.6} & \crcell{49.0}{32.6}{48.8}{32.4} & \crcell{47.1}{37.6}{47.5}{33.1} & \crcell{48.9}{43.5}{41.6}{31.3} & \val{49.86}{3.45} \\
\bottomrule
\end{tabular}
\caption{Per quantization $\times$ sampling-strategy estimates of \textbf{accuracy (\%)} for Qwen 2.5 VL 7B and Gemma 4 E4B. Each video cell reports Constrained (C) and Reasoning (R) settings as mean\,$\pm$\,std; image-based datasets are frame-independent and span all sampling columns.}
\label{tab:acc_only}
\end{table*}

\section{Captioning Quality Tests}~\label{app:caption-quality}

Because the captioner $C$ is held frozen across every method, we tested their sensitivity before committing to Nvidia's Nemotron 3 Nano Omni. Table~\ref{tab:caption-quality-overall} shows that across six viable candidates, CLIPScore on frames randomly sampled from the six tested long-video datasets spanned only 0.2331 to 0.2447 CLIPScore, a range of roughly 5\%. Importantly, only Gemma 4 E4B scored particularly worse at 0.2153. Since modern VLMs are capable of captioning images, this result is not quite surprising.

This brings us to the proposition of efficiency. To make LazySloth fast, we need $C$ to balance captioning quality with latency. Across the six candidates, we note that the per-caption latency spanned 0.59 to 2.70 seconds, demonstrating a difference of 4.6$\times$ from the fastest to the slowest model. Since $C$ is invoked once per tree node and therefore sits in the inner loop of every method we compare, latency is the dimension that actually separates the candidates. Nemotron 3 Nano Omni was the fastest captioner that was not a quality outlier. Therefore, we chose this model for LazySloth and across all methods so that no baseline is advantaged or penalized. This allowed us to cleanly compare the merit of each retrieval method.

Moreover, looking at Table~\ref{tab:caption-quality-perdataset}, we want to discuss Nemotron's high CLIPScore despite the fact that it wrote the second-shortest captions. As seen in Table~\ref{tab:caption-quality-overall}, CLIPScore is directly correlated to average caption length (a known property of the metric). This makes Nemotron's per-word caption quality competitive. Second, the per-dataset spread within any model is comparable to the spread between models, which means the differences in Table~\ref{tab:caption-quality-perdataset} do not account for a strict ranking.

\begin{table}[t]
\centering
\scriptsize
\setlength{\tabcolsep}{4pt}
\begin{tabular}{@{}l cccc@{}}
\toprule
\textbf{Model} & \makecell{CLIP\\Score $\uparrow$} & \makecell{SigLIP\\Score $\uparrow$}
& \makecell{Avg.\ length\\(words)} & \makecell{Latency\\(s) $\downarrow$} \\
\midrule
google/gemma-4-12b            & \textbf{0.2447} & 0.1367 & 32.6 & 1.27 \\
qwen/qwen3.6-27b              & 0.2435 & \textbf{0.1403} & 31.4 & 2.29 \\
google/gemma-4-31b            & 0.2417 & 0.1348 & 29.1 & 2.70 \\
qwen/qwen3.5-9b               & 0.2417 & 0.1402 & 31.5 & 0.92 \\
qwen2.5-vl-7b-instruct        & 0.2341 & 0.1292 & 27.7 & 0.98 \\
\rowcolor{cyan!10}
nvidia/nemotron-3-nano-omni   & 0.2331 & 0.1340 & 25.1 & \textbf{0.59} \\
google/gemma-4-e4b-it         & 0.2153 & 0.1175 & 21.8 & 0.77 \\
\bottomrule
\end{tabular}
\caption{Reference-free caption quality of the candidate frozen captioners $C$,
pooled over all evaluated datasets. Captions were generated from the same sampled
frames with the identical pipeline prompt, then scored by CLIPScore
\citep{hessel-etal-2021-clipscore} and a SigLIP2 cross-check. Latency is
wall-clock seconds per caption. The shaded row is the captioner used in all
experiments reported in this paper.}
\label{tab:caption-quality-overall}
\end{table}

\begin{table}[t]
\centering
\scriptsize
\setlength{\tabcolsep}{3pt}
\begin{tabular}{@{}l ccccc@{}}
\toprule
\textbf{Model} & \makecell{Ego\\Schema} & \makecell{LongVideo\\Bench}
& \makecell{MMBench\\Video} & \makecell{Movie\\Chat} & \makecell{Video\\MME} \\
\midrule
google/gemma-4-31b            & 0.236 & 0.239 & 0.246 & 0.248 & 0.238 \\
qwen/qwen3.6-27b              & 0.236 & 0.238 & 0.245 & 0.262 & 0.237 \\
google/gemma-4-12b            & 0.227 & 0.239 & 0.248 & 0.264 & 0.245 \\
qwen/qwen3.5-9b               & 0.236 & 0.229 & 0.244 & 0.259 & 0.240 \\
qwen2.5-vl-7b-instruct        & 0.229 & 0.230 & 0.230 & 0.256 & 0.226 \\
\rowcolor{cyan!10}
nvidia/nemotron-3-nano-omni   & 0.234 & 0.220 & 0.230 & 0.252 & 0.230 \\
google/gemma-4-e4b-it         & 0.181 & 0.219 & 0.220 & 0.229 & 0.229 \\
\bottomrule
\end{tabular}
\caption{Per-dataset CLIPScore for each candidate captioner. Rows are ordered by
the pooled CLIPScore of Table~\ref{tab:caption-quality-overall}.}
\label{tab:caption-quality-perdataset}
\end{table}

\section{Additional Efficiency Analysis of LazySloth}
\label{app:efficiency-analysis}

For a fine-grained understanding of the efficiency analysis in Section~\ref{sec:eff-analysis}, we broke down median inference time under each method and direct inference in Table~\ref{tab:runtime_comparison} for the benchmark with the longest videos, LVBench. We found that inference time under each retrieval-based method grew linearly with video length, but the informative quantity is the slope of growth. While LazySloth paid 2.43 additional seconds per extra minute of video (s/min), VideoLucy paid 5.53s, and WorldMM paid 19.86s/min. We note that these differences were significant, as WorldMM paid 17.43 s/min more than LazySloth ($t = 13.2$, $p < 10^{-39}$) and VideoLucy 3.10 s/min more ($t = 4.9$, $p < 10^{-5}$). LazySloth therefore addressed the scaling problem that these methods struggled with.

This difference, however, stemmed from both the number of caption calls and how they were structured. As shown in Table~\ref{tab:caption_calls}, LazySloth's caption count was effectively compared to WorldMM and the brute-forced baseline, growing only from 35 on 40-minute-long videos to 45 under 140-minute-long videos due to the $O(k \log_k N)$ bound of Equation~(1). WorldMM grew from 97 to 330, and the brute-forced eager trees from 3,600 to 12,603, a factor of 280 at the longest videos. VideoLucy, however, is the interesting case as it issued 27 to 54 calls, statistically indistinguishable from ours, yet still ran roughly three$\times$ slower. 

This is because VideoLucy re-captions dense fixed-length overlapping windows, so each call carries more frames and more visual tokens than a LazySloth bunch caption. Figure~\ref{fig:decomposition} shows that captioning consumed 67\% of its total time. Our bounded search cost precisely addressed this issue.

\begin{table*}[t]
\centering
\begin{tabular}{lccccc}
\toprule
\textbf{Length} & \textbf{LazySloth} & \textbf{VideoLucy} & \textbf{WorldMM} & \textbf{Brute-forced Eager} & \textbf{DI} \\
\midrule
40 min  & 149 s & 509 s (3.4$\times$) & 735 s (4.9$\times$) & 7.0 h (169$\times$)  & 57 s \\
60 min  & 176 s & 553 s (3.1$\times$) & 1109 s (6.3$\times$) & 10.5 h (214$\times$) & 57 s \\
90 min  & 251 s & 727 s (2.9$\times$) & 1560 s (6.2$\times$) & 15.8 h (226$\times$) & 57 s \\
120 min & 287 s & 1000 s (3.5$\times$) & 2393 s (8.3$\times$) & 21.0 h (263$\times$) & 57 s \\
140 min & 307 s & 936 s (3.0$\times$) & 2371 s (7.7$\times$) & 24.5 h (287$\times$) & 57 s \\
\bottomrule
\end{tabular}
\caption{Median inference time comparison across different video lengths on LVBench. Speedup is shown relative to LazySloth.}
\label{tab:runtime_comparison}
\end{table*}

\begin{table*}[t]
\centering
\begin{tabular}{lcccc}
\toprule
\textbf{Length} & \textbf{LazySloth} & \textbf{VideoLucy} & \textbf{WorldMM} & \textbf{Brute-forced Eager} \\
\midrule
40 min  & 35 & 27 (0.8$\times$) & 97 (2.8$\times$)  & 3,600 (103$\times$) \\
60 min  & 35 & 35 (1.0$\times$) & 149 (4.3$\times$) & 5,401 (156$\times$) \\
90 min  & 38 & 46 (1.2$\times$) & 220 (5.8$\times$) & 8,102 (215$\times$) \\
120 min & 34 & 54 (1.6$\times$) & 291 (8.6$\times$) & 10,801 (320$\times$) \\
140 min & 45 & 42 (0.9$\times$) & 330 (7.3$\times$) & 12,603 (280$\times$) \\
\bottomrule
\end{tabular}
\caption{Median caption calls across different video lengths on LVBench. Speedup is shown relative to LazySloth with Qwen3.6 27B.}
\label{tab:caption_calls}
\end{table*}

\begin{figure}[ht]
    \centering
    \includegraphics[width=\columnwidth]{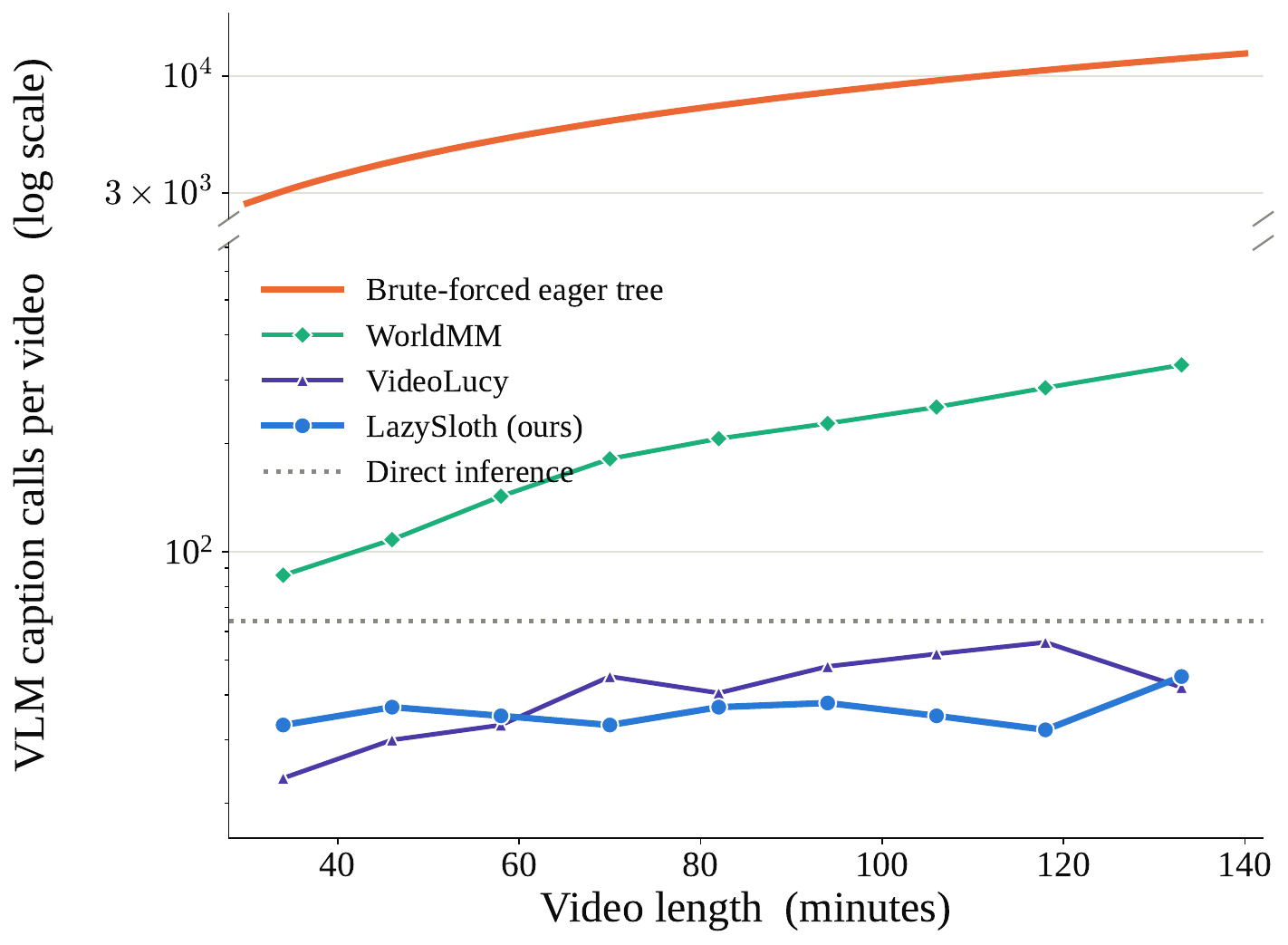}
    \caption{WorldMM vs VideoLucy vs LazySloth median captions generated per-video on LVBench. While VideoLucy doesn't caption significantly more than LazySloth, the difference in wall clock time arises from how the two methods utilize caching and optimize multi-frame inputs to VLMs.}
    \label{fig:caption}
\end{figure}

\section{Additional Benchmark-wise Behavioral Analysis}

\begin{figure*}
    \centering
    \includegraphics[width=\textwidth]{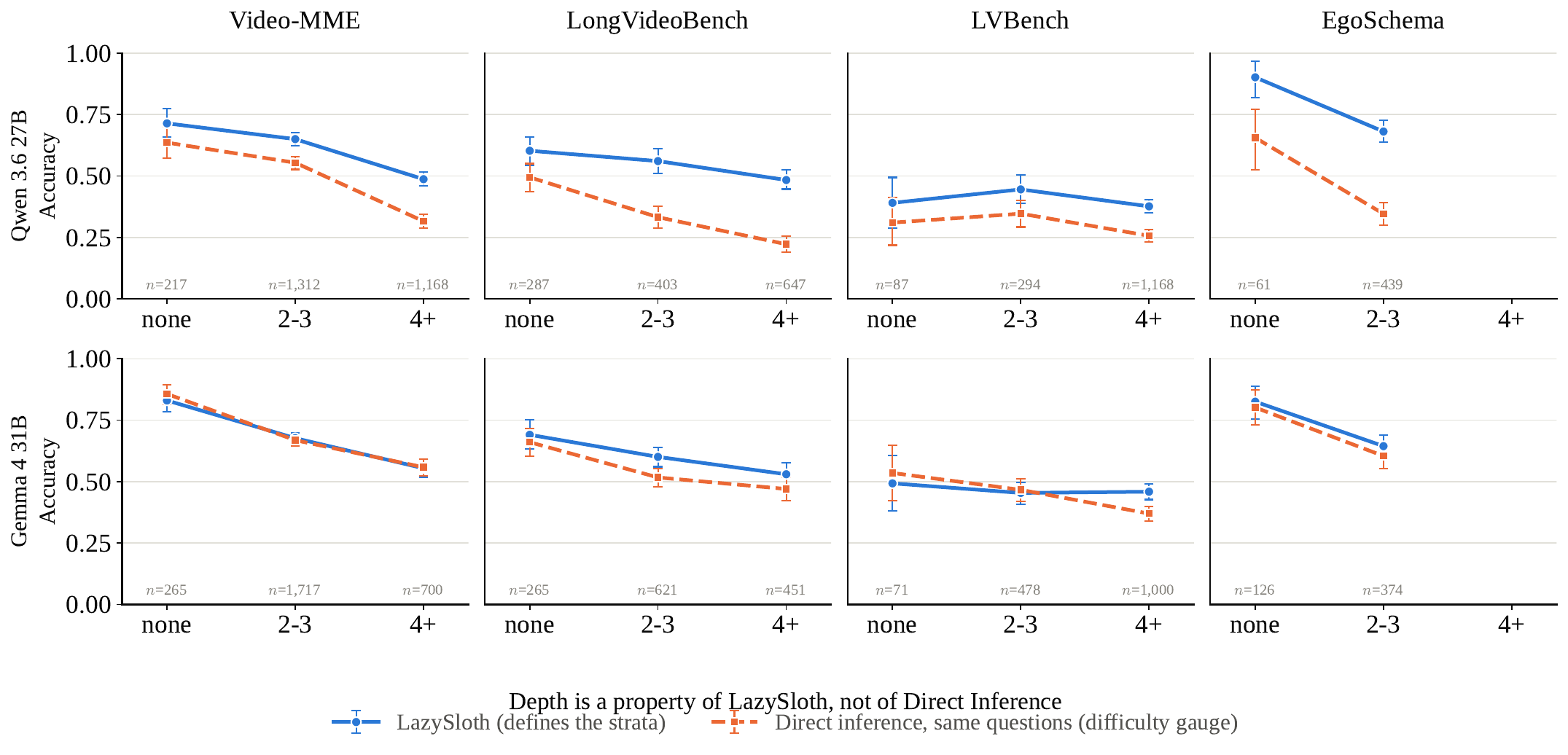}
    \caption{Tree depth searched by the two tested base VLMs, Gemma 4 31B and Qwen3.6 27B, across the four datasets with LazySloth and Direct Inference. Since Direct Inference does not create trees to search in, it is scored on exactly the questions that fall in each of LazySloth's strata.}
    \label{fig:depth-by-dataset}
\end{figure*}

To better understand the impact of tree depth searched by the searcher $S$ on final accuracy, we plotted the aggregate depth trend of Figure~\ref{fig:behavior-depth} by benchmark and backbone in Figure~\ref{fig:depth-by-dataset}. Direct inference has no search and therefore no depth of its own; it is scored on exactly the questions that fall in each of LazySloth's strata. The decline we plotted in Figure~\ref{fig:depth-by-dataset} against Direct Inference indicated whether the deeper-searched strata are indeed the harder questions. This helps us eliminate an important confound that could misguide this discussion. The takeaway, therefore, is the performance difference between direct inference and LazySloth, which we plotted with 95\% confidence interval bars.

We note that for Qwen3.6 27B, LazySloth consistently improved performance from direct inference, with the differences reaching statistical significance at higher tree depths across the four benchmarks, most sharply on EgoSchema and LongVideoBench. 

On Gemma 4 31B, however, the two curves were essentially coincident on EgoSchema and LVBench, and LazySloth directionally improved performance on LongVideoBench, although not statistically significant. Since both the blue and orange lines in the plot refer to identical questions within each stratum, the difference cannot be attributed to question difficulty, and instead reflects how much a given base model gains from our method. Gemma 4 31B consistently struggled to gain any performance across the board, supporting our claim that gains are model-dependent, and Gemma 4 31B lacks any headroom to improve performance with retrieval-based or agentic frameworks.

\begin{figure}[t]
    \centering
    \includegraphics[width=0.9\columnwidth]{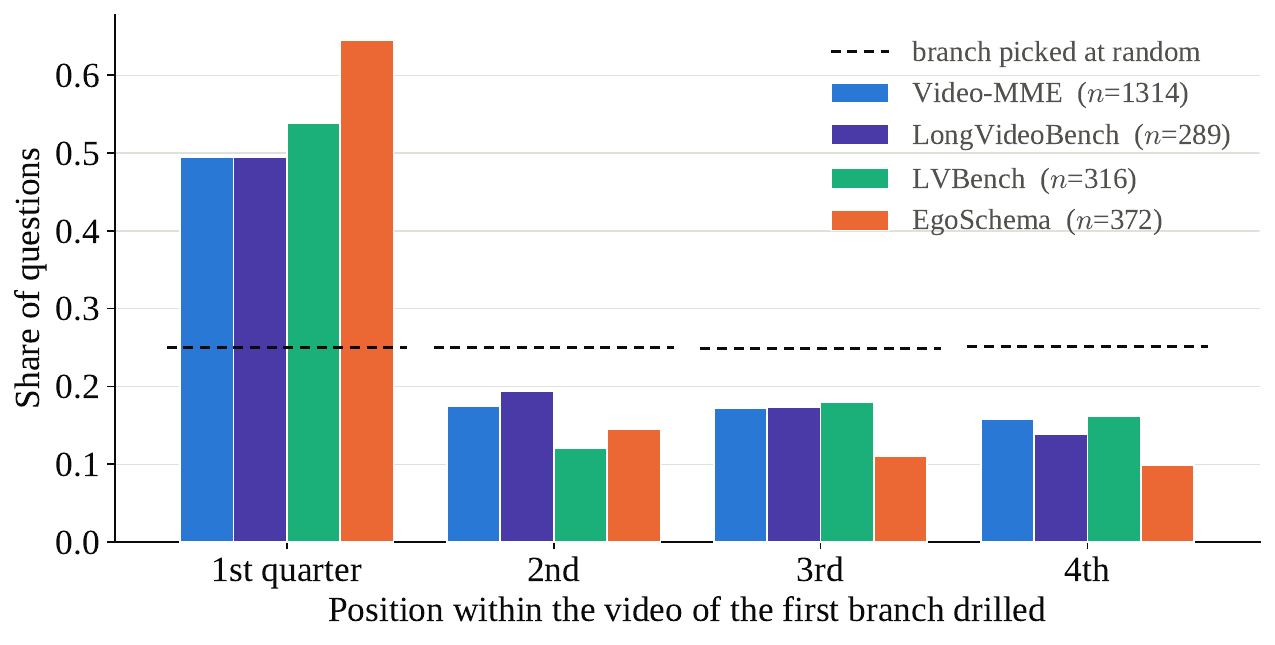}
    \caption{Share of questions for which LazySloth (aggregated over Qwen3.6 27B and Gemma 4 31B) chose the first, second, third, or fourth quarter of the video at the first level of the search tree. LazySloth's first descent was strongly biased toward the start of the video (roughly twice the rate of random branch selection). The dashed reference shows an empirically measured random baseline.}
    \label{fig:behavior-pos-bias}
\end{figure}

Further, Figure~\ref{fig:behavior-pos-bias} examines which branch the search descends into first against randomly picking any branch from the generated children. We note that models preferred the first frontier split consistently more than the other three branches, which have a roughly equal chance of being picked. The first descent landed in the opening quarter of the video on 49\% of Video-MME and LongVideoBench questions, 54\% of LVBench, and 64\% of EgoSchema, which was roughly twice the random rate.

Finally, Table~\ref{tab:behavior-qatype} matched LazySloth against direct inference on the same questions and split the result by the question type. We note that gains in performance were not uniform. Question types, mostly those requiring a moment to be located and then reasoned about, produced the largest improvements (LongVideoBench's T2O gained 44.7\%, SOS 38.3\%, and TOS 37.0\%, and Video-MME's Information Synopsis and Temporal Reasoning gained 25.5\% and 23.2\% respectively). This is an architectural improvement LazySloth brings to the table. Types answerable from any representative frame, as expected, gained almost nothing. Our method helps the most in cases where finding the right moment is the task, which explains the performance differences reported in Table~\ref{tab:main-results}.

\begin{table*}
\centering
\small
\begin{tabular}{lrrrrrr}
\toprule
Benchmark & Question type & $N$ & LazySloth & Direct inf. & $\Delta$ & McNemar $\chi^2$ \\
\midrule
LongVideoBench & SSS & 97 & 38.1 & 12.4 & +25.8 & 18.6$^{*}$ \\
LongVideoBench & E3E & 94 & 52.1 & 28.7 & +23.4 & 13.8$^{*}$ \\
LongVideoBench & S2E & 93 & 57.0 & 43.0 & +14.0 & 4.4$^{*}$ \\
LongVideoBench & S2A & 88 & 64.8 & 62.5 & +2.3 & 0.0 \\
LongVideoBench & O2E & 87 & 50.6 & 42.5 & +8.0 & 1.9 \\
LongVideoBench & TAA & 82 & 46.3 & 26.8 & +19.5 & 7.0$^{*}$ \\
LongVideoBench & SOS & 81 & 61.7 & 23.5 & +38.3 & 23.1$^{*}$ \\
LongVideoBench & T2A & 79 & 54.4 & 51.9 & +2.5 & 0.0 \\
LongVideoBench & T2O & 76 & 65.8 & 21.1 & +44.7 & 28.7$^{*}$ \\
LongVideoBench & T3O & 74 & 50.0 & 14.9 & +35.1 & 22.3$^{*}$ \\
LongVideoBench & TOS & 73 & 47.9 & 11.0 & +37.0 & 20.5$^{*}$ \\
LongVideoBench & T3E & 73 & 43.8 & 24.7 & +19.2 & 7.0$^{*}$ \\
LongVideoBench & SAA & 72 & 58.3 & 27.8 & +30.6 & 17.0$^{*}$ \\
LongVideoBench & S2O & 72 & 61.1 & 40.3 & +20.8 & 7.3$^{*}$ \\
LongVideoBench & O3O & 66 & 54.5 & 28.8 & +25.8 & 8.8$^{*}$ \\
LongVideoBench & E2O & 65 & 47.7 & 43.1 & +4.6 & 0.2 \\
LongVideoBench & T2E & 65 & 52.3 & 27.7 & +24.6 & 9.4$^{*}$ \\
Video-MME & Object Reasoning & 453 & 57.8 & 36.0 & +21.9 & 55.5$^{*}$ \\
Video-MME & Object Recognition & 354 & 65.3 & 60.5 & +4.8 & 2.5 \\
Video-MME & Information Synopsis & 322 & 72.4 & 46.9 & +25.5 & 54.7$^{*}$ \\
Video-MME & Action Recognition & 313 & 55.0 & 52.7 & +2.2 & 0.4 \\
Video-MME & Action Reasoning & 285 & 54.7 & 33.3 & +21.4 & 33.0$^{*}$ \\
Video-MME & Counting Problem & 267 & 37.1 & 32.2 & +4.9 & 1.6 \\
Video-MME & Attribute Perception & 222 & 67.6 & 63.5 & +4.1 & 1.0 \\
Video-MME & Temporal Reasoning & 177 & 42.4 & 19.2 & +23.2 & 20.8$^{*}$ \\
Video-MME & OCR Problems & 139 & 56.8 & 61.2 & -4.3 & 0.6 \\
Video-MME & Spatial Reasoning & 56 & 76.8 & 58.9 & +17.9 & 4.0$^{*}$ \\
Video-MME & Temporal Perception & 55 & 70.9 & 61.8 & +9.1 & 0.8 \\
Video-MME & Spatial Perception & 54 & 70.4 & 61.1 & +9.3 & 1.1 \\
\bottomrule
\end{tabular}
\caption{Search gain by question type using Qwen 3.6 27B as the base VLM. LVBench and EgoSchema are omitted as they do not contain segregated question types. $^{*}$ marks $\chi^2 > 3.84$ ($p < 0.05$).}
\label{tab:behavior-qatype}
\end{table*}

\begin{table*}[t]
\centering
\footnotesize
\resizebox{\textwidth}{!}{%
\begin{tabular}{l p{11cm}}
\toprule
\textbf{Symbol} & \textbf{Meaning} \\
\midrule
$V$ & Input video clip. \\
$Q$ & Query/task (MCQ question + options, open-ended question, or classification prompt). \\
$L$ & Label set --- MCQ letters / class labels; $\varnothing$ for open-ended QA (free-text answer). \\
$C$ & Frozen captioner VLM; only ever produces captions (kept fixed so an experiment varies only the search VLM). \\
$S$ & Search VLM + reasoner VLM --- one model in two roles: text-only navigation, then multimodal final answer. \\
$\phi$ & Frame sampling rate, measured in FPS, used to decode the clip. \\
$F = \langle f_1, \dots, f_N \rangle$ & Ordered extracted frames; each $f_i = (\text{idx}_i, t_i, \text{path}_i)$ = native index, timestamp, JPEG path. \\
$N$ & Number of extracted frames, $= \lvert F \rvert$. \\
$k$ & Branch factor --- children per \textsc{Drill}, and the root's fan-out. \\
$\ell$ & Leaf size --- frame-range threshold; once a node spans $\le \ell$ frames, \textsc{Drill} stops and \textsc{Expand} (per-frame captions) is allowed. \\
$m_b$ & Max representative frames shown to $C$ when captioning one node/bunch (evenly subsampled); ``b'' = bunch. \\
$m_r$ & Max key frames shown to the reasoner $S$ (evenly subsampled, time order); ``r'' = reasoner. \\
$R$ & Max VLM search rounds (round budget). \\
$s_{\max}$ & Consecutive no-progress rounds before an answer is forced ($=3$). \\
$\mathcal{F}$ (frontier) & The currently-shown sibling nodes --- the candidates for \textsc{Drill}/\textsc{Expand}. \\
prev / $\mathcal{F}_{\text{prev}}$ & The previous frontier, retained so $S$ can \textsc{Drill} a sibling to backtrack one level. \\
$\mathcal{E}$ (evidence) & The caption corpus surfaced along the path (root $\to$ branch $\to$ per-frame captions). \\
$[a,b)$ / \texttt{currPos} & The position range into $F$ the traversal narrowed to; the reasoner's key-frame window (starts $[0, N)$). \\
\texttt{node.pos} $=[a_c, b_c)$ & A node's half-open position range into $F$; its span $b_c - a_c$ is compared against $\ell$. \\
level / id & Node depth (root $=0$) / node id (its start frame index --- stable, unique within a level). \\
$\text{cap}[\,\cdot\,]$ & On-disk caption cache, keyed by the frozen captioner $C$ (semantic tree for nodes, episodic for frames); reused across queries. \\
$y_{\text{ag}}, y_{\text{re}}, \hat{y}$ & The VLM's answer, the reasoner's answer, and the final prediction ($\hat{y} = y_{\text{re}}$, or $y_{\text{ag}}$ if the reasoner returns \textsc{Insufficient}). \\
\bottomrule
\end{tabular}%
}
\caption{Notation reference for formalization of the LazySloth framework in Section~\ref{sec:ssa-pipeline} and Algorithm~\ref{alg:main}.}
\label{tab:notation}
\end{table*}

\end{document}